\documentclass[runningheads]{llncs}
\usepackage{tikz}
\usetikzlibrary{positioning,shapes,arrows}
\tikzset{
    events/.style={ellipse, draw, align=center},
}
\usepackage{graphicx}
\usepackage{xcolor}
\usepackage{subcaption}
\usepackage{enumitem}
\usepackage{amsmath}
\usepackage{booktabs}
\usepackage{algorithm}
\usepackage{algorithmic}
\usepackage{multirow}
\newcommand{\comment}[1]{}
\newcommand{\FC}[1]{\textcolor{blue}{(Fahima): #1}}
\newcommand{\hlFC}[1]{\setlength{\fboxsep}{0pt}\colorbox{cyan}{#1}}

\begin{document}
\title{A Symbolic Approach for Counterfactual Explanations \\ (Preprint version)}
%
%
\author{Ryma Boumazouza \orcidID{0000-0002-3940-8578}
 \and
Fahima Cheikh-Alili \orcidID{0000-0002-4543-625X} 
\and
Bertrand Mazure \orcidID{0000-0002-3508-123X}
 \and
Karim Tabia \orcidID{0000-0002-8632-3980}
}
\authorrunning{R. Boumazouza et al.}
%
\institute{CRIL, Univ. Artois and CNRS, F62300 Lens, France
\{boumazouza,cheikh,mazure,tabia\}@cril.univ-artois.fr
}
\maketitle              
\begin{abstract}
We propose a novel symbolic approach to provide counterfactual explanations for a classifier predictions. Contrary to most explanation approaches where the goal is to understand which and to what extent parts of the data helped to give a prediction, counterfactual explanations indicate which features must be changed in the data in order to change this classifier prediction. Our approach is symbolic in the sense that it is based on encoding the decision function of a classifier in an equivalent CNF formula. In this approach, counterfactual explanations are seen as the Minimal Correction Subsets (MCS), a well-known concept in knowledge base reparation. Hence, this approach takes advantage of the strengths of already existing and proven solutions for the generation of MCS. Our preliminary experimental studies on Bayesian classifiers show the potential of this approach on several datasets.

\end{abstract}
\section{Introduction}
Recently, a symbolic approach for explaining classifiers has been proposed in \cite{shih2018symbolic}. This approach first compiles a classifier  into an {\bf equivalent} and  {\bf tractable} symbolic representation then enumerates some forms of explanations such as prime implicants.  It has many nice features in terms of tractability, explanation enumeration and formal analysis of classifiers. 
This paper proposes a novel approach that is designed to equip such symbolic approaches \cite{shih2018symbolic} with a module for counterfactual explainability. Intuitively, we view the process of computing counterfactual explanations as the one of computing Minimal Correction Subsets (MCS generation) where the knowledge base stands for the classifier and the data instance in hand. As we will show later, our symbolic approach for counterfactual generation has many nice features added to the fact of lying on well-known concepts and efficient existing techniques for MCS generation. \\
The inputs to our approach are a classifier's decision function $f$ compiled into an equivalent symbolic representation in the form of an Ordered Decision Diagram ($ODD$) and a data instance $x$. Our contribution is to model the problem of counterfactual generation as the one of MCS generation. We will show the properties of this encoding and highlight the links between MCS and counterfactual explanations. Our experiments show that using existing MCS generation tools, one can efficiently compute counterfactual explanations as far as a classifier can be compiled into an ODD which is the case of Bayesian network classifiers \cite{shih2018symbolic}, decision trees and some neural nets \cite{shi2020tractable}.

\section{From a symbolic representation of a classifier to an equivalent CNF encoding}
Our approach for counterfactual explanation proceeds in two steps : The first one is encoding a symbolic representation (given in the form of an ODD) into an equivalent CNF representation. The second step consists in computing MCSs meant as counterfactual explanations given the CNF representation of the classifier and any data instance. In this section, we describe the first step of our approach. 
Let us first formally recall some definitions used in the remainder of this paper. For the sake of simplicity, the presentation is limited to binary classifiers with binary features 
but the approach still applies to non binary classifiers as stressed in \cite{shih2018symbolic}.

\begin{definition}{\textbf{(Binary Classifier)}} A Binary Classifier is defined by two sets of variables: A feature space $X$= \{$X_1$,...,$X_n$\} where $|X|=n$, and a binary class variable denoted $Y$. Both the features and the class variable take values $in$ \{0,1\}.
\end{definition}
 
 \begin{definition}{\textbf{(Decision Function of A Classifier)}} A decision function of a classifier ($X,Y$) is a function $f:X \rightarrow Y$ mapping each instantiation  $x$ of $X$ to $y$=$f(x)$.
 \end{definition}
A decision function describes the classifier's behavior independently from the way it is implemented. 
\begin{definition}{(\textbf{Ordered  Decision Diagram ODD}}) \label{Defodd}
An Ordered  Decision Diagram (ODD) is a rooted, directed acyclic graph, defined over an ordered set of discrete variables, and encoding a decision function. Each node is labeled with a variable $X_i$ $, i= 1,\dots,n$ and has an outgoing edge corresponding to each value $x_i$ of the variable $X_i$, except for the sink nodes, which represent the terminal nodes. 
\end{definition}
An Ordered Binary Decision Diagram OBDD is an ODD where all the variables are binary.
If there is an edge from a node labeled $X_i$ to a node labeled $X_j$ , then $i$$<$$j$ (more on tractable representations such as ODDs can be found in \cite{shih2019compiling}). 

\begin{example}\label{exBNODD}
Figure \ref{fig:adm_bnc} shows a naive Bayes classifier for deciding whether a student will be admitted to a university (class variable: Admit (A)). The features of an applicant are: work-experience (WE),
first-time-applicant (FA),
entrance-exam (E)
and gpa (GPA).
In Figure \ref{fig:adm_odd}, we provide the OBDD representing the classifier decision function $f$ with the variable ordering (WE, FA, E, GPA). Here, the sinks correspond to the values of the class variable (A). \vspace{-0.9cm}\\
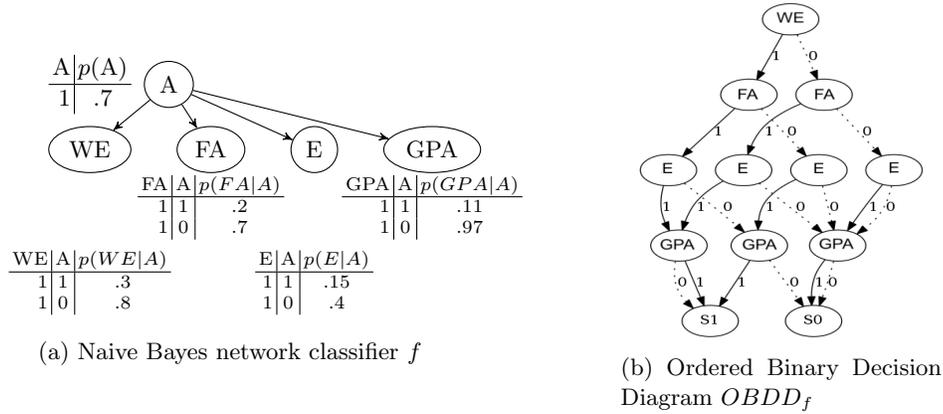
\begin{figure}[h]
\begin{subfigure}{0.5\textwidth}
\centering
\begin{tikzpicture}[node distance=.6cm, >=stealth']
\node [events] (A) {\small A};
\node [events, below left = of A] (WE) {\small WE};
\node [events, right = of WE] (FA) {\small FA};
\node [events, right = of FA] (E) {\small E};
\node [events, right = of E] (GPA) {\small GPA};

\draw [->] (A) -- (WE);
\draw [->] (A) -- (FA);
\draw [->] (A) -- (E);
\draw [->] (A) -- (GPA);

\node [left =.05cm of A, anchor=east] {\small
\begin{tabular}{r|c} 
A & $p$(A)  \\ \hline
1 & .7 \\
\end{tabular}
};

\node [below =2cm of WE, anchor=south] {\scriptsize
\begin{tabular}{r|c|c} 
WE & A & $p(WE|A)$  \\ \hline
1 & 1 &.3 \\
1 & 0 &.8 \\
\end{tabular}
};

\node [below =1cm of FA, anchor=south] {\scriptsize
\begin{tabular}{r|c|c} 
FA & A & $p(FA|A)$  \\ \hline
1 & 1 &.2 \\
1 & 0 &.7 \\
\end{tabular}
};

\node [below =2cm of E, anchor=south] {\scriptsize
\begin{tabular}{r|c|c} 
E & A & $p(E|A)$  \\ \hline
1 & 1 &.15 \\
1 & 0 &.4 \\
\end{tabular}
};

\node [below =1cm of GPA, anchor=south] {\scriptsize
\begin{tabular}{r|c|c} 
GPA & A & $p(GPA|A)$  \\ \hline
1 & 1 &.11 \\
1 & 0 &.97 \\
\end{tabular}
};



\end{tikzpicture}
\caption{Naive Bayes network classifier $f$}
\label{fig:adm_bnc}
\end{subfigure}
\hspace{2cm}
\begin{subfigure}{0.35\textwidth}
\centering
\includegraphics[width=0.9\linewidth, height=4.5cm]{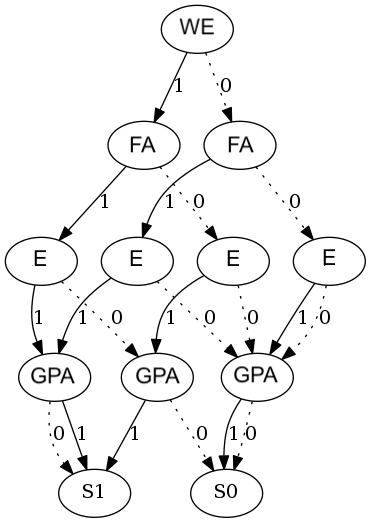}
\caption{Ordered Binary Decision Diagram $OBDD_f$}
\label{fig:adm_odd}
\end{subfigure}
\caption{A Naive Bayes network classifier and its corresponding OBDD.}
\label{step1}
\end{figure}

\end{example}\vspace{-.9cm}
\comment{
During the compilation process, the ODD variables are numbered after integers ranging from 1 to N(total number of nodes of the ODD). A node of level $i$ corresponds to the $i^{th}$ feature in the variable ordering of the ODD. \\
For example in Fig.\ref{step1}, given $M$, its equivalent \hlFC{ODD$_M$} \FC{$OBDD_M$. Pour les termes contenant des indices, il faut avoir toute la notation entre \$\$.} and the variable ordering (WE, FA, E, GPA), the node (numbered "0") of the $1^{st}$ level refers to the variable "WE", those of the $2^{nd}$ level ("1" and "8") to the variable "FA". \FC{Il faut compléter l'exemple en précisant ce que signifie $WE$, $FA$,... et quelle est la décision à prendre (que veut dire "admit"?).}\\
}

Let us now focus on our target representation. A CNF (Clausal Normal Form)  formula is a conjunction  of clauses. A clause is a formula composed of a disjunction of literals. A literal is either a Boolean variable or its negation. A quantifier-free formula is built from atomic formulae using conjunction $\wedge$, disjunction $\vee$, and negation $\neg$. An interpretation $\mu$ assigns values from \{0, 1\} to every Boolean variable. Let $\Sigma$ be a CNF formula, $\mu$ satisfies  $\Sigma$ iff $\mu$ satisfies all clauses of $\Sigma$.\\
There are several methods to encode a decision diagram as a CNF formula. For instance in \cite{cabodi2003improving}, the authors proposed a method called "Single-Cut-Node" to store a BDD (Binary Decision Diagram) as a CNF where they  model BDD nodes as multiplexers. 
A second method called "The No-Cut method" creates clauses starting from $f$ corresponding to the “off-set” and a last method called "The Auxiliary-Variable-Cut" which combines the two other methods. For the sake of simplicity and clarity, we choose the simplest method which does not involve adding new variables during the encoding process 
since we want to restrict our explanations to the input variables of the classifier. We implement a simple way to encode the symbolic representation of a classifier as a CNF formula based on the "The No-Cut" method \cite{cabodi2003improving}. 
In our case, since we are dealing with binary Boolean functions (binary features and class variable), our tractable representation of the decision function $f$ is an OBDD. We use along with this paper positive/true/1 and negative/false/0 interchangeably. Let us first define an "off-set" of a Boolean function and a CNF formula.

\begin{definition}{\textbf{(Off-Set of a Boolean function)}}\label{offsetdef}
The Off-set of a Boolean function $f$, denoted as $f^0$, is $f^0$=$\{v \in \bigcup^{n}_{i=1} \{0,1\}^{i}  | f(v)=0\}$
If $f^0$=$\{0,1\}^n$, then $f$ is unsatisfiable. Otherwise, $f$ is satisfiable. 
\end{definition}
Intuitively, $f^0$ is the set of counter-models of $f$. 
This concept of "off-set" contains the counter-models we need to enumerate in order to construct our CNF's clauses. 
The OBDD is used to enumerate all the paths from the root to the $0$-sink node (the off-set), where each element of $f^0$ corresponds to a path within it. 
\comment{

\begin{example}[Example \ref{exBNODD} cont'd]
Let us continue with the OBDD of Example \ref{exBNODD}. 
The corresponding "off-set" is: $f^0$=\{(1,2,-3,-4), (1,-2,3,-4), (1,-2,-3), (-1,2,-3), (-1,-2)\}, representing the counter models of the OBDD. The notation "-2" means that the variable numbered 2 is instantiated negatively (has the value 0). The element (1,-2,3,-4) represents the instantiation $x_1$=1 $\wedge $ $x_2$=0 $\wedge$ $x_3$=1 $\wedge$ $x_4$=0.  Note that the elements of $f^0$ are a complete assignation but are written in a reduced form for the sake of simplicity. 
\end{example}
}

\begin{definition} \label{odd2cnf}(\textbf{CNF encoding of an OBDD})
Let $f$ be the decision function encoded by an ordered binary decision diagram $OBDD_f$. Let also Off-set($OBDD_f$) be the off-set of $OBDD_f$. We define the obtained CNF formula from $OBDD_f$ as $\Sigma_{f}$=$\wedge$$\neg$$e_i$ where $e_i$$\in$$f^{0}$ and $i$=$1$..$|f^{0}|$.
\end{definition}
Let $\alpha$ be the associated formula of $f$. The intuition is that $\neg$ $\alpha$ $\equiv$ $\vee$ $e_i$ where $e_i$ $\in$ $f^{0}$. Then $f$ comes down to negating $\vee$ $e_i$ allowing to obtain directly $f$ in the form of a CNF.
\comment{

}  
Following Definition \ref{odd2cnf}, we have:
\begin{itemize}
    \item[-] Every variable of the feature space  $X$= \{$X_1$,...,$X_n$\} of the classifier will correspond to a Boolean variable in the CNF $\Sigma_{f}$.
    \item[-] The class variable $Y$ of the classifier is captured by the truth value of the CNF ($\Sigma_{f}$). 
    \item[-] Modeling a prediction made by the classifier for a given data instance $x$ comes down to the truth value of: (CNF $\Sigma_{f}$ $\wedge$ $\Sigma_x$) where $\Sigma_x$ stands for the data instance $x$ encoded as a CNF by a set of unit\footnote{A unit clause involves only one Boolean variable represented by a literal} clauses.
\end{itemize}

\comment{\begin{example}
Let $f$ be the decision function of the classifier represented by an OBDD in figure \ref{xorobdd}.
The running example shows the CNF $\Sigma_{f}$ corresponding to the $OBDD_f$. The CNF clauses in \ref{offsetCNFfCNFx}(a) are generated over the "off-set" of $f$ within the OBDD. The figure \ref{offsetCNFfCNFx}(b) illustrate how we represent the CNF $\Sigma_x$ of the corresponding data instance $x=(1,0)$.

\begin{figure}[ht]
\begin{minipage}[b]{0.45\linewidth}
\centering
     \begin{center}
            \begin{tabular}{c c c c}
             p & cnf & 2 & 2 \\ 
              1 & 2 & 0 \\  
             -1 & -2 & 0 
           \end{tabular}
            \caption*{(a): CNF $\Sigma_{f}$ }
     \end{center}
\label{cnfFM}
\end{minipage}
\hspace{0.5cm}
\vline
\begin{minipage}[b]{0.45\linewidth}
\centering
    \begin{center}
            \begin{tabular}{c c c c}
             p & cnf & 2 & 2 \\ 
             1 & 0 \\  
             -2 & 0 
           \end{tabular}
           \caption*{(b): CNF $\Sigma_x$}
           
    \end{center}

\label{cnfX}
\end{minipage}
\caption{The CNF Boolean formula associated to the decision function $f$ represented in the example \ref{xorobdd} and the data instance $x=(1,0)$.}
\label{offsetCNFfCNFx}
\end{figure}
\end{example}
}

Our encoding guarantees the logical equivalence between the $OBDD_f$ and the obtained CNF $\Sigma_{f}$. The following proposition formally states this result. 

\begin{proposition}\label{theoremcnf}
Let $f$ be a binary decision function and $OBDD_f$ its compiled representation.  Let also $\Sigma_{f}$ be the CNF representation of the decision function $f$ obtained following Definition \ref{odd2cnf}. Then an interpretation $\mu$ is model of $\Sigma_{f}$ iff it is mapped to 1 by $f$.  
\end{proposition}
Proposition \ref{theoremcnf} states that $\Sigma_f$ is {\it logically equivalent} to the function $f$. i.e, they have the same truth value for each data  instance $x$. Thus, we can assert the following result. 
\begin{lemma}\label{lemm2}
Given a binary classifier, a data instance $x$ and the predicted class $f(x)=y$, ($\Sigma_{f}$ $\wedge$ $\Sigma_{x}$) is {\it SAT}  iff $f(x)$=$1$ . 
\end{lemma}

We stress that both of the compilation of the classifier to a symbolic representation and the encoding into a CNF formula, is done only once in our approach and can be re-used to explain as many instances as wanted.

\comment{
In this section, we are given the decision function $f$ of a binary classifier in the form of an OBDD and we provide its equivalent CNF encoding denoted $\Sigma_f$ and its main properties. The next section presents how we generate counterfactual explanation for a given data instance $x$.
} 

\section{Generating Counterfactual Explanations}\label{counterfactXAI}
Intuitively, a counterfactual explanation for an instance of interest $x$ and a classifier $f$ is the minimum changes to $x$ needed to alter the output of $f$. 
Let us now define formally the concept of counterfactual explanation.
\begin{definition}{\textbf{(Counterfactual Explanation)}}\label{ce}
Let $x$ be a complete data instance and $f(x)$ its prediction by the decision function $f$. A counterfactual explanation $\acute{x}$ of $x$ is such that:
\begin{itemize}
    \item $\acute{x} \subseteq x$ ($\acute{x}$ is a subset or a part of x)
    \item $f(x[\acute{x}])$= 1-$f(x)$ ( prediction inversion)
    \item There is no $\hat{{x}}$ $\subset$ $\acute{x}$ such that $f(x[\hat{{x}}])$=$f(x[\acute{x}])$ (minimality)
\end{itemize}
\end{definition}
In Definition \ref{ce}, the term $x[\acute{x}]$ denotes the data instance $x$ where variables included in $\acute{x}$ are reversed.
In our setting, a counterfactual $\acute{x}$ is defined as a part of the data instance $x$ such that $\acute{x}$ is minimal and $\acute{x}$ allows to flip the prediction $f(x)$. 
The explanation comes as follow: an output $y$ is returned because variables from the features space $X$ had values ($x_1, x_2...$). If instead, $X$ had values ($x'_1, x'_2, ...$) while all other variables had remained constant, the output $y'$ would have been returned. 
Counterfactual explanations are expected to explain both the outcome of a prediction and how that would change if {\it things had been different}. In an "approved vs rejected" application like the common example of loan applications, a counterfactual explanation would answer the question {\it What do I need to change in my application for the bank to approve my application ?}.
Our main idea is to model the counterfactual explanation task as a Partial Max-SAT problem. Recall that our CNF encoding of an OBDD representation of a classifier's decision function $f$ ensures that a negative (resp. positive) prediction  leads to an unsatisfiable (resp. satisfiable) CNF Boolean formula. Namely, $f(x)$=1 iff ($\Sigma_f$$\wedge$$\Sigma_x$) is satisfiable. In the case where $f(x)$=0, $\Sigma_f$ $\wedge$ $\Sigma_x$ is unsatisfiable and it is possible to identify the subsets of $\Sigma_x$ allowing to restore the consistency of $\Sigma_f$$\wedge$$\Sigma_x$ (recall that $\Sigma_f$ is satisfiable unless the classifier $f$ always predicts $0$ regardless of the instance $x$). This is a well-known problem dealt with in many areas such as knowledge base reparation, consistency restoration, etc. We will see later how to provide explanations when the outcome is positive, namely $f(x)$=1, using the same mechanisms. 
 It is important to note that our CNF is composed of two parts: $\Sigma_f$ and $\Sigma_x$, where $\Sigma_f$ encodes the classifier and it is satisfiable and $\Sigma_x$ encode the data  instances $x$ and represented as a set of unit clauses.
In the case of negative predictions, since $\Sigma_f$$\wedge$$\Sigma_x$ is unsatisfiable (inconsistent), we can compute a sort of reparation set that is composed of the subsets of data instance $x$ that cause the unsatisfiability of  $\Sigma_f$$\wedge$$\Sigma_x$. This is known as the minimal correction subset (MCS). 
\begin{definition}{\textbf{(MSS)}}
A maximal satisfiable subset (in short, MSS) $\Phi$ of a CNF $\Sigma$ is a subset (of clauses) $\Phi$ $\subseteq$ $\Sigma$ that is satisfiable and such that $\forall$ $\alpha$ $\in$ $\Sigma$ $\backslash$ $\Phi$, $\Phi$ $\cup$ \{$\alpha$\} is unsatisfiable.
\end{definition}

\begin{definition}{\textbf{(MCS (Co-MSS))}}
A minimal correction subset (in short
MCS, also called Co-MSS) $\Psi$ of a CNF $\Sigma$ is a set of formulas $\Psi$ $\subseteq$ $\Sigma$ whose complement in $\Sigma$, i.e., $\Sigma$ $\backslash$ $\Psi$, is an MSS of $\Sigma$
\end{definition}
In our case, given a data instance $x$ and a function $f$, and their respective CNFs $\Sigma_x$ and $\Sigma_f$, an MCS $\Phi$ ensures the minimality property and tells what clauses to remove from $\Sigma_f$$\wedge$$\Sigma_x$ to restore its consistency. 
Note that although the number of MCSs can be exponential in the worst case, it remains low in many benchmarks. 

\comment{
\begin{example}
Let $\Sigma$ be an unsatisfiable CNF formed by a set
of clauses \{$\alpha1$, $\alpha2$, $\alpha3$, $\alpha4$, $\alpha5$, $\alpha6$\}, where $\alpha1$ = $a$, $\alpha2$ = $b$ , $\alpha3$ = $c$, $\alpha4$ = $\neg a$  $\vee$ $\neg b$, $\alpha5$ = $\neg a$ $\vee$ $\neg c$, $\alpha6$ = $\neg b$ $\vee$ $\neg c$. The MCSs of $\Sigma$ are \{$\alpha1$, $\alpha3$\}, \{$\alpha2$, $\alpha3$\} and \{$\alpha1$, $\alpha2$\}. \newline
\end{example}
}
\subsection{Counterfactuals for  negative predictions through MCSs}
Our main idea for explaining a negative prediction for an instance $x$ is to compute its MCSs. An MCS identifies the subset of clauses to be repaired to restore the satisfiability of the CNF formula. In order for an MCS to correspond to a counterfactual explanation, it should contain only the unit clauses belonging to $\Sigma_x$ indicating what features need to be removed or flipped such that the whole CNF (namely, $\Sigma_f$$\wedge$$\Sigma_x$) becomes again satisfiable. This leads to splitting the CNF into two subsets: hard constraints (those that could not be included in any MCS) and soft ones (those that could be relaxed, hence included in MCSs). 
The concepts of Partial Max-SAT, Hard and Soft constraints(clauses) are 
defined as follows:
\comment{
\begin{definition} (\textbf{Max-SAT})
Given a Boolean CNF formula $\Sigma$, the Max-SAT is the problem of finding a truth assignment that satisfies the maximum number of clauses in $\Sigma$. 
\end{definition}
}
\begin{definition}(\textbf{Partial Max-SAT})
Given a Boolean CNF formula $\Sigma$ in which some clauses are hard and some are soft, Partial Max-SAT is the problem of finding a truth assignment that satisfies all the hard constraints and the maximum number of soft ones.
\end{definition}
In order to solve Partial Max-SAT, we will consider the general setting,
where a formula is composed of two disjoint sets of clauses $\Sigma$ = $\Sigma_H$ $\cup$ $\Sigma_S$ \cite{biere2009handbook}, where  $\Sigma_H$ denotes the hard clauses and $\Sigma_S$ denotes the soft ones.
\comment{
\begin{definition} (\textbf{Hard and Soft clauses})
Let $\Sigma_1$ and $\Sigma_2$ be two sets of clauses where $\Sigma_2$ is satisfiable. Partial Max-SAT( $\Sigma_1$,$\Sigma_2$) computes one cardinality maximal subset of  $\Sigma_1$ that is satisfiable with  $\Sigma_2$.  $\Sigma_1$ and $\Sigma_2$ are called the sets of soft and hard constraints (clauses), respectively.
\end{definition}
}
In our modeling for counterfactual generation, the set of hard clauses is $\Sigma_f$ while soft clauses is $\Sigma_x$ representing the data instance $x$ to explain. The CNF encoding of the classifier $\Sigma_f$ as a set of hard clauses is presented in the previous section. The CNF encoding of the data instance $\Sigma_x$ as soft clauses is done as follows. Let $\Sigma_x$ be the {\it soft Clauses}, defined as follow: 
\begin{itemize}
    \item Each clause $\alpha \in \Sigma_x$ is composed of exactly one literal ($\forall \alpha \in \Sigma_x, |\alpha|= 1$)
    \item Each literal representing a Boolean variable of $\Sigma_x$ corresponds to a Boolean variable $\{ X_i \in X / i \in [1,n]\}$ of the feature space of the decision function $f$.
    
\end{itemize}

Following our approach, an MCS for $\Sigma_f$$\wedge$$\Sigma_x$ comes down to a subset of soft clauses, namely a part of $x$ that is enough to remove in order to restore the consistency, hence to flip the prediction $f(x)$=0. Proposition \ref{cemsc} states that each MCS  computed for $\Sigma_f$$\wedge$$\Sigma_x$ represents a counterfactual explanation $\acute{x}\subseteq x$ for the prediction $f(x)$=0 and vice versa.

\begin{proposition}\label{cemsc}
Let $f$ be the decision function, $OBDD_f$ its compiled symbolic representation, $x$ be a data instance predicted negatively ($f(x)=0$) and $\Sigma_f$$\wedge$$\Sigma_x$ an unsatisfiable CNF.  
Let $CF(x,f)$ 
be the set of counterfactuals of $x$ wrt. $f$.
Let MCS$(\Sigma_{f,x})$ the set of MCSs of $\Sigma_f\wedge\Sigma_x$.\\
Then: 
\begin{equation}\label{mcsiscfiff}
        \forall \acute{x} \subseteq x, \acute{x} \in CF_f(x,f(x))   \iff \acute{x} \in MCS(\Sigma_{f,x})
\end{equation}

\end{proposition}

The MCS enumeration is done over the {\it soft clauses}, which practically should reduce the time needed to enumerate all the MCS since we will have less clauses to consider. 
As for positively predicted instances, we can simply work on the negation of $OBDD_f$ representation of the decision function $f$ namely, we will rely on the CNF $\Sigma_{\neg f}$$\wedge$$\Sigma_x$ to compute the counterfactuals in a similar way. 
\section{Experiments}
This section provides our experiments to evaluate  our approach of generating counterfactual explanations. Given a data instance of interest $x$ and the decision function $f$, we encode both $f$ and $x$ as a CNF formula and model counterfactuals generation as a {\it Partial maximum satisfiability problem} formed by conjoining  $\Sigma_{f}$ and $\Sigma_x$.
We start by considering typical binary classification problems and focus on {\it Binary Naive Bayes Classifiers (BNC)}. To test our approach, we compiled synthetic naive Bayes classifiers to OBDDs using the approach proposed in \cite{shih2019compiling}, then we encode these OBDDs into CNF formulas before getting into the generation of the MCSs. We mention that for each network size, we used an average of 4 to 10 networks to run our experiments. The decision threshold used for the compilation of the synthetic BNCs is $.5$. 
\subsection{Compiling Bayes classifiers into OBDDs} 
 Table \ref{tab1:bnc-sdd}  summarizes the compilation experiments we ran on BNCs with different sizes. For each category of classifiers having the same number of features, we compute the average size of their corresponding OBDD (number of nodes). 
\begin{table}[htb!]
\centering\vspace{-.5cm}
        \begin{tabular}{|r|p{1cm}|p{1cm}|p{1cm}|p{1cm}|p{1cm}|p{1cm}|}
        \hline
         Nb\_Features & 5 & 10 & 16 & 20 & 22 & 25\\
         \hline
         OBDD$_{size}$ & 9 & 42 & 370 & 1020 & 2546 & 8626\\  
         \hline
       \end{tabular}
       \caption{Average size of the OBDD representations.
       }\label{tab1:bnc-sdd}
\end{table}\vspace{-1cm}
As expected, we notice a large increase in the OBDD size as the number of features of the classifier grows up. Moreover, it seems that the OBDD size also strongly depends on the classifier's parameters in addition to the number of features.
\subsection{Dumping OBDDs as CNF formulas}
Next step will be to encode the obtained OBDD into CNF Boolean formulas, simply denoted CNF($f$). 
The latter has the same variables as the classifier. 
We aim in the following to compare the size of both OBDDs and CNFs of classifiers with different sizes. 
\comment{
\begin{table}[h!]
\begin{center}\vspace{-.5cm}
        \begin{tabular}{|r|p{1cm}|p{1cm}|p{2cm}|p{2cm}|}
         \hline
         OBDD$_{size}$ & 12 & 66 &  1429 & 8857\\  
         \hline
         CNF$_{size}$ & 6 & 116 & 2164 & 717781
           \\ \hline 
         Avg\_Runtime (ms)  & 1.4 & 32.3 & 241806.5 & 327888500
           \\ \hline 
       \end{tabular}
       \caption{ Average size and runtime required to encode OBDDs into CNFs  }
       \label{tab1:graph-cnf}
\end{center}\vspace{-1cm}
\end{table}
}
As observed experimentally in Table \ref{tab1:nbmcs}, the time and size (number of clauses)  of the generated CNF are strongly correlated to the size of its OBDD. 
While the compilation time scales linearly with the number of nodes of the OBDD, the size of the CNF can be much smaller, depending on the classifier’s parameters (threshold of the used BNC and variable order used for OBDD). We remind that this encoding is done only once for a given classifier and, then, can subsequently be used for explaining any number of instances.

\subsection{MCSs generation}
Once we got the CNF ($\Sigma_{f} \wedge \Sigma_{x}$), we can get to the generation of MCSs. 
In our experiments, we use the boosting algorithm for the MCSs generation proposed in \cite{gregoire2018boosting} and implemented in \textbf{{ \it EnumELSRMRCache\footnotemark[1]: a tool for MCSs enumeration}}. \footnotetext[1]{available at \url{http://www.cril.univ-artois.fr/enumcs/}}
Recall that our input is a CNF composed of hard clauses (encoding the classifier) and soft ones (encoding the data instance to explain). 
The data instances were randomly generated. 

\begin{table}[h!]
\begin{center}\vspace{-0.5cm}
        \begin{tabular}{|r|p{1cm}|p{1cm}|p{1cm}|p{1cm}|p{1cm}|p{1.5cm}|}
        \hline
         $\#$Vars & 5 & 10 & 16 & 20 & 22 & 25\\  
         \hline
         OBDD$_{size}$ & 9 & 42 & 370 & 1020 & 2546 & 8626\\  
         \hline
         CNF$_{size}$  & 3 & 64 & 2598 & 27122 & 123878 & 684847\\
         \hline
         Encoding\_Runtime (ms)  & 1.4 & 32.3 & 2725 & 241806 & 430471 & 327888500
           \\ \hline 
          $\#$MCS & 3 & 23 & 101 & 305 & 272  & 364 \\
         \hline
         Runtime ${(ms)}$ & 1.9 & 2.3 & 17.8 & 1762 & 9299.4 & 109148.5 \\
          \hline
        
       \end{tabular}
       \caption{ Average size of OBDD/CNF,  runtime (ms), and number of the counterfactual explanations (MCSs). 
       }
       \label{tab1:nbmcs}
\end{center}\vspace{-1cm}
\end{table}

The aim here is to compare the number of counterfactual explanations with the size of OBDD and CNF representations.
Table \ref{tab1:nbmcs} summarizes the results of the average number of counterfactual explanations generated given a data instance and a classifier. The experiments are carried out on classifiers, OBDDs and CNFs with different sizes. 
As expected, the number of explanations increases with the CNF size in general, but remains strongly related to: (1) the classifier and OBDD parameters (variable ordering), and (2) to the data instance itself. 
As shown in Table \ref{tab1:nbmcs}, the average run-time does not seem to depend on the number of MCSs generated but more on the number of features of the classifier, which is expected since the time-consuming part of generating the MCSs is related to the size of the representations in terms of number of clauses and their size.\\
To sum up the results, it can be said that  as long as we can get a symbolic tractable representation of a classifier, our approach can provide counterfactual explanations.  
The number of different MCSs of the CNF Boolean formulas
remains low in our case since our approach 
computes the MCSs over the soft clauses only, which experimentally significantly reduces the time of MCSs enumeration. 
Finally, the obtained results suggest that the number of counterfactuals is probably very small compared with other types of explanations computed for symbolic representations (e.g. prime-implicant, minimum cardinality, etc,.), but this remains to be confirmed on benchmarks with different properties.
\section{Concluding remarks}
 The approach proposed in this paper allows to equip the symbolic approach proposed in \cite{shih2019compiling} with a module for counterfactual explanations. Our approach is simple and takes advantage of well-defined concepts and proven tools for MCSs. Moreover, our approach is specifically designed to provide actionable explanations. 
The main issue currently is that it requires a compilation step to get the symbolic representation of a classifier (compilers already exist for Bayesian networks \cite{shih2019compiling}, decision trees and some neural nets \cite{shi2020tractable}).
Another problem that need to be treated is the scaling
of the compilation algorithm for classifiers
with a large number of features. \\
\bibliographystyle{splncs04}
\bibliography{sumbib}

\begin{thebibliography}{1}
\providecommand{\url}[1]{\texttt{#1}}
\providecommand{\urlprefix}{URL }
\providecommand{\doi}[1]{https://doi.org/#1}

\bibitem{biere2009handbook}
Biere, A., Heule, M., van Maaren, H.: Handbook of satisfiability, vol.~185. IOS
  press (2009)

\bibitem{cabodi2003improving}
Cabodi, G., Nocco, S., Quer, S.: Improving sat-based bounded model checking by
  means of bdd-based approximate traversals. In: 2003 Design, Automation and
  Test in Europe Conference and Exhibition. pp. 898--903. IEEE (2003)

\bibitem{gregoire2018boosting}
Gr{\'e}goire, {\'E}., Izza, Y., Lagniez, J.M.: Boosting mcses enumeration. In:
  IJCAI. pp. 1309--1315 (2018)

\bibitem{shi2020tractable}
Shi, W., Shih, A., Darwiche, A., Choi, A.: On tractable representations of
  binary neural networks. arXiv preprint arXiv:2004.02082  (2020)

\bibitem{shih2018symbolic}
Shih, A., Choi, A., Darwiche, A.: A symbolic approach to explaining bayesian
  network classifiers. arXiv preprint arXiv:1805.03364  (2018)

\bibitem{shih2019compiling}
Shih, A., Choi, A., Darwiche, A.: Compiling bayesian network classifiers into
  decision graphs. In: Proceedings of the AAAI Conference on Artificial
  Intelligence. vol.~33, pp. 7966--7974 (2019)

\end{thebibliography}
\end{document}